\documentclass{article}

\usepackage{arxiv}

\usepackage[utf8]{inputenc} 
\usepackage[T1]{fontenc}    
\usepackage{hyperref}       
\usepackage{url}            
\usepackage{booktabs}       
\usepackage{multirow}
\usepackage{amsmath}
\usepackage{amsfonts}       
\usepackage{nicefrac}       
\usepackage{microtype}      
\usepackage{graphicx}
\graphicspath{ {./images/} }

\title{FRWKV: FREQUENCY-DOMAIN LINEAR ATTENTION FOR LONG-TERM TIME SERIES FORECASTING}

\author{
 Qingyuan Yang$^{1,*}$ \\
  College of Information Science and Engineering\\
  Northeastern University\\
  \texttt{2400951@stu.neu.edu.cn} \\
  \And
 Shizhuo Deng$^{1,2,\dagger}$ \\
  College of Information Science and Engineering\\
  Northeastern University\\
  \texttt{dengshizhuo@mail.neu.edu.cn} \\
  \And
 Dongyue Chen$^{1,2,\ddagger}$ \\
  College of Information Science and Engineering\\
  Northeastern University\\
  \texttt{chendongyue@ise.neu.edu.cn} \\
  \And
 Da Teng$^{1}$ \\
  College of Information Science and Engineering\\
  Northeastern University\\
  \texttt{13166672732@163.com} \\
  \And
 Zehua Gan$^{1}$ \\
  College of Information Science and Engineering\\
  Northeastern University\\
  \texttt{2370888@stu.neu.edu.cn} \\
}

\begin{document}
\maketitle
\begin{abstract}
Traditional Transformers face a major bottleneck in long-sequence time series forecasting due to their quadratic complexity $(\mathcal{O}(T^2))$ and their limited ability to effectively exploit frequency-domain information. Inspired by RWKV's $\mathcal{O}(T)$ linear attention and frequency-domain modeling, we propose FRWKV, a frequency-domain linear-attention framework that overcomes these limitations. Our model integrates linear attention mechanisms with frequency-domain analysis, achieving $\mathcal{O}(T)$ computational complexity in the attention path while exploiting spectral information to enhance temporal feature representations for scalable long-sequence modeling. Across eight real-world datasets, FRWKV achieves a first-place average rank. Our ablation studies confirm the critical roles of both the linear attention and frequency-encoder components. This work demonstrates the powerful synergy between linear attention and frequency analysis, establishing a new paradigm for scalable time series modeling. Code is available at this repository: \url{https://github.com/yangqingyuan-byte/FRWKV}.
\end{abstract}


\section{INTRODUCTION}
Long-term time series forecasting (LTSF) is a critical task with significant applications across various fields, including energy dispatching, transportation management, finance, and meteorology \cite{wolff2020gaussian,li2024dynamic,thompson2016multifractal}. Accurate prediction of future trends from historical data is essential for informed decision-making and strategic planning. The field has seen a progression from traditional statistical models like ARIMA \cite{box2015time} to advanced deep learning methods, including CNNs \cite{borovykh2017conditional}, RNNs \cite{lin1996learning,oreshkin2019n,qn2017dual,salinas2020deeper}, and, more recently, Transformer-based architectures.

The advent of Transformer-based models, such as PatchTST \cite{nie2023time}, TimesNet \cite{wu2023timesnet}, and iTransformer \cite{liu2024itransformer}, has marked a significant breakthrough in LTSF. These models have demonstrated superior capabilities in capturing complex temporal dependencies. However, their reliance on the traditional self-attention mechanism, which has a quadratic time and memory complexity of $\mathcal{O}(T^2)$ with respect to the sequence length $T$, poses a major bottleneck. This limitation makes them computationally prohibitive for long sequences and hinders their scalability.

To address the computational challenges of full-attention mechanisms, linear attention mechanisms have been proposed. These methods, including Performer \cite{choromanski2020rethinking} and RWKV \cite{peng2023rwkv,peng2025rwkv7,hou2024rwkvts}, achieve a more favorable linear complexity $(\mathcal{O}(T))$, enabling them to scale effectively to long sequences. Despite their efficiency, these models primarily operate in the time domain, which may limit their ability to effectively exploit frequency-domain information and capture periodic patterns inherent in many time series datasets.

To address the limitations of purely time-domain approaches, a separate line of research has focused on integrating frequency-domain knowledge into deep learning architectures. Frequency-domain methods are powerful tools for capturing multi-scale and periodic patterns. Models like TimesNet \cite{wu2023timesnet}, FreEformer \cite{yue2025freeformer} and FreTS \cite{yi2024frequency} leverage this approach to enhance feature representations. However, while these models improve predictive accuracy, they often do not fundamentally optimize the underlying computational complexity. Consequently, they still face significant computational and memory challenges when handling long sequences. This is evidenced by TimesNet's performance, which suffers significantly in both training and inference under such conditions. This gap highlights a clear opportunity for a model that combines the efficiency of linear attention with the robustness of frequency-domain analysis.

We propose FRWKV, a novel framework that combines linear attention with frequency-domain analysis. FRWKV adapts linear attention mechanisms to operate in the frequency domain. By mapping sequences to spectra and executing state recursion and gating directly in the frequency domain, our model achieves a linear attention path with $\mathcal{O}(T)$ complexity. This approach not only allows it to scale seamlessly to long inputs but also leverages frequency-domain priors to stabilize the extraction of periodic and resonant structures.

Our model addresses the aforementioned challenges by: (i) overcoming the quadratic complexity of traditional attention, (ii) enhancing the expressiveness of linear attention by operating in the frequency domain, and (iii) effectively utilizing frequency-domain information for more robust and accurate long-term forecasting.

Across standard power, transportation, weather, and solar benchmarks, FRWKV delivers competitive or state-of-the-art results, with pronounced gains at long terms. Ablation studies confirm that the frequency-domain formulation consistently improves robustness to cross-scale periodicity and non-stationarity.

We make the following contributions:
\begin{itemize}
    \item \textbf{Frequency-domain linear attention architecture:} We instantiate linear attention in the frequency domain, providing a simple, stable, and scalable linear-attention alternative for LTSF.
    \item \textbf{Scalability with linear complexity:} The attention path remains $\mathcal{O}(T)$, supporting long inputs within practical memory budgets.
    \item \textbf{Empirical evidence under long terms:} Extensive experiments across diverse domains show consistent benefits at long prediction horizons; code and hyperparameters are released to ensure reproducibility.
\end{itemize}

\begin{figure}[t]
    \centering
    \includegraphics[width=0.9\linewidth]{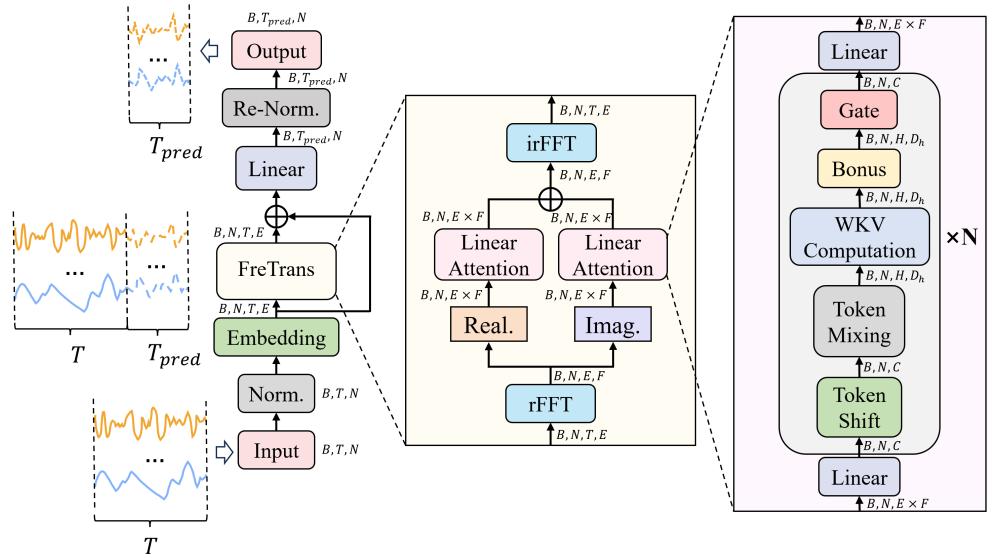}
    \caption{FRWKV architecture: The input time series undergoes normalization and embedding. The embedding is transformed to real and imaginary components by real-valued Fast Fourier Transform (rFFT). The real and imaginary components are processed separately through linear attention mechanisms, then concatenated and transformed back to the time domain via inverse real-valued Fast Fourier Transform (irFFT) and normalization to yield the final output.}
\end{figure}

\section{METHODOLOGY}
\subsection{Problem Definition}
Given a multivariate time series $\mathbf{X} = [\mathbf{x}_1,\mathbf{x}_2,\dots ,\mathbf{x}_T]\in \mathbb{R}^{N\times T}$, where $N$ is the number of variables and $T$ is the sequence length, the long-term forecasting task aims to predict future values $\mathbf{Y} = [\mathbf{x}_{T + 1},\mathbf{x}_{T + 2},\ldots ,\mathbf{x}_{T + \tau}]\in \mathbb{R}^{N\times \tau}$ for a horizon $\tau$. Our goal is to learn a mapping function $f:\mathbb{R}^{N\times T}\to \mathbb{R}^{N\times \tau}$ that minimizes the prediction error while maintaining computational efficiency.

\subsection{Model Overview}
We use the following symbols in the paper and Fig.~1: $B$ denotes batch size; $T$ is the input sequence length (time steps); $N$ is the number of variables; $E$ is the embedding dimension; $C$ is the model dimension (d\_model); $F = \lfloor T / 2\rfloor +1$ is the number of frequency bins after rFFT; $H$ is the number of attention heads and $D_{h}$ is the attention head size, satisfying $H\cdot D_h = C$.

FRWKV is an end-to-end encoder-decoder that performs linear attention in the frequency domain (Fig.~1). Inputs are first normalized by RevIN~\cite{kim2022reversible} and converted to complex spectra with rFFT in FreTrans module, which are split into real/imaginary branches. Each branch is encoded by a linear-attention module (RWKV-style state recursion, $\mathcal{O}(T)$), then recomposed by irFFT and projected to the horizon; RevIN finally restores the original scale.

\subsection{Frequency-Domain Transformation}
For a real-valued sequence $x \in \mathbb{R}^T$, the real FFT (rFFT) maps it to the $F = \lfloor T / 2 \rfloor + 1$ non-redundant frequency bins:
\begin{equation}
X _ {\mathrm {r}} [ k ] = \sum_ {n = 0} ^ {T - 1} x [ n ] e ^ {- j 2 \pi k n / T}, \quad k = 0, \dots , F - 1.
\end{equation}

The inverse rFFT (irFFT) reconstructs the real signal by completing the spectrum and applying the inverse sum:
\begin{equation}
x [ n ] = \frac {1}{T} \sum_ {k = 0} ^ {T - 1} X [ k ] e ^ {j 2 \pi k n / T}, \quad 0 \leq n \leq T - 1,
\end{equation}
where
\begin{equation}
X [ k ] = \begin{cases}
X _ {\mathrm {r}} [ k ], & 0 \leq k \leq F - 1, \\
\overline {{X _ {\mathrm {r}} [ T - k ]}}, & F \leq k \leq T - 1.
\end{cases}
\end{equation}

In long-horizon forecasting, we apply rFFT along the temporal axis of multivariate inputs $\mathbf{X} \in \mathbb{R}^{B \times N \times T}$ to obtain compact spectra $\mathbf{X}_{\mathrm{freq}} \in \mathbb{C}^{B \times N \times F}$. The real and imaginary parts are encoded separately by our frequency branches. For each branch, we reshape the spectra to $\mathbf{Z} \in \mathbb{R}^{T \times C}$ by flattening the batch and variable dimensions, and irFFT brings features back to the time domain for horizon-wise projection.

\subsection{Linear Attention in Frequency Encoders}
The state recursion performs selective forgetting along the current-key direction and writes new information via the value-key outer product. We denote the per-branch token sequence by $\mathbf{Z} = [\mathbf{z}_1, \mathbf{z}_2, \dots, \mathbf{z}_T] \in \mathbb{R}^{T \times C}$, where each token $\mathbf{z}_t \in \mathbb{R}^C$ represents the projected features at time step $t$. The encoder implements linear-time attention through three essential steps that correspond to Fig.~1.

Per time step $t$: $\mathbf{z}_t \in \mathbb{R}^C$ is the current token, $\tilde{\mathbf{z}}_t \in \mathbb{R}^C$ is the shifted token, and $\bar{\mathbf{z}}_t \in \mathbb{R}^C$ is the mixed token. $\pmb{\mu} \in [0,1]^C$ is a learnable channel-wise mixing coefficient. Linear maps yield $\mathbf{r}_t, \mathbf{k}_t, \mathbf{v}_t \in \mathbb{R}^C$, which are then reshaped into $H$ heads with dimensions $\mathbb{R}^{H \times D_h}$ (where $H \cdot D_h = C$). The gate $\mathbf{g}_t \in (0,1)^C$ modulates the output. Lightweight MLPs produce per-channel decay $\mathbf{d}_t \in (0,1)^C$ and replacement strength $\mathbf{i}_t \in (0,1)^C$. We denote the per-head recurrent state and transition by $\mathbf{S}_t$, $\mathbf{G}_t \in \mathbb{R}^{H \times D_h \times D_h}$. Keys are normalized/reweighted as $\tilde{\mathbf{k}}_t$ (per-head $\ell_2$ normalization) and $\hat{\mathbf{k}}_t$ (data-driven replacement using $\mathbf{i}_t$). The bonus multiplier $\mathbf{B} \in \mathbb{R}^{C \times C}$ is diagonal and $\mathbf{W}_o \in \mathbb{R}^{C \times C}$ is the output projection; the layer output is $\mathbf{o}_t \in \mathbb{R}^C$.

\paragraph{Token shift and mixing.} A shift operator $S$ augments local dynamics. Let $\tilde{\mathbf{z}}_t = S(\mathbf{z})_t = \mathbf{z}_{t-1}$ with $\tilde{\mathbf{z}}_1 = 0$. The mixed token is computed as:
\begin{equation}
\bar {\mathbf {z}} _ {t} = (1 - \boldsymbol {\mu}) \odot \mathbf {z} _ {t} + \boldsymbol {\mu} \odot \tilde {\mathbf {z}} _ {t}, \quad \boldsymbol {\mu} \in [ 0, 1 ] ^ {C},
\end{equation}
then apply linear maps (omitted for brevity) to obtain $\mathbf{r}_t, \mathbf{k}_t, \mathbf{v}_t$ and a gate $\mathbf{g}_t \in (0,1)^C$. Two lightweight MLPs produce per-channel decay $\mathbf{d}_t \in (0,1)^C$ and replacement strength $\mathbf{i}_t \in (0,1)^C$.

\paragraph{State-recursive update.} Let $\mathbf{S}_t \in \mathbb{R}^{H \times D_h \times D_h}$ be the per-head recurrent state. With a normalized key $\hat{\mathbf{k}}_t$ we define
\begin{equation}
\mathbf {G} _ {t} = \operatorname {Diag} \left(\mathbf {d} _ {t}\right) - \tilde {\mathbf {k}} _ {t} \mathbf {i} _ {t} ^ {\top},
\end{equation}
where $\mathbf{d}_t, \tilde{\mathbf{k}}_t, \mathbf{i}_t \in \mathbb{R}^{H \times D_h}$ are per-head tensors (reshaped from $\mathbb{R}^C$), and update/output as
\begin{align}
\mathbf {S} _ {t} &= \mathbf {G} _ {t} \mathbf {S} _ {t - 1} + \mathbf {v} _ {t} \hat {\mathbf {k}} _ {t} ^ {\top}, \\
\mathbf {y} _ {t} &= \mathbf {S} _ {t} \mathbf {r} _ {t}.
\end{align}
These recursions avoid token-token interactions and thus yield an attention path of $\mathcal{O}(T)$ (contrast with quadratic Transformer attention).

\paragraph{Bonus coupling and gating.} We enhance receptance-replacement interaction via
\begin{align}
\beta_ {t} &= \mathbf {r} _ {t} ^ {\top} \mathbf {B} \hat {\mathbf {k}} _ {t}, \\
\mathbf {o} _ {t} &= \mathbf {g} _ {t} \odot \mathbf {W} _ {o} (\mathbf {y} _ {t} + \beta_ {t} \mathbf {v} _ {t}),
\end{align}
where $\mathbf{B}$ is a learnable diagonal matrix. The output $\mathbf{o}_t$ feeds the next layer. At each time step, these updates process a length-$T$ sequence in a single forward scan with linear-time complexity.

\section{EXPERIMENTS}
\subsection{Experimental Setup}
We evaluate our FRWKV on eight real-world datasets spanning power, transportation, weather and solar. Dataset statistics are in Table~\ref{tab:dataset}. Following prior work, we use input length $T = 96$ and horizons $\{96, 192, 336, 720\}$. Full experimental details are in open-source code.

\begin{table}[t]
    \centering
    \caption{Dataset details.}
    \label{tab:dataset}
    \begin{tabular}{lcc}
        \toprule
        Datasets & Features & Timestamps \\
        \midrule
        ETTm1 & 7 & 69680 \\
        ETTm2 & 7 & 69680 \\
        ETTh1 & 7 & 17420 \\
        ETTh2 & 7 & 17420 \\
        ECL & 321 & 26304 \\
        Exchange & 9 & 7588 \\
        Weather & 21 & 52696 \\
        Solar & 137 & 52179 \\
        \bottomrule
    \end{tabular}
\end{table}

\subsection{Main Results}
Table~\ref{tab:main} shows FRWKV attains the best average rank (2.00 MSE, 1.38 MAE) across 8 datasets, with frequent 1st/2nd places. It maintains competitive performance on long horizons while preserving linear $\mathcal{O}(T)$ attention complexity, indicating robust long-range modeling across power, traffic, weather, and finance.

\begin{table*}[t]
    \centering
    \caption{Main results averaged over $T=96$ and $\tau\in\{96,192,336,720\}$. Ranks in parentheses; smaller is better.}
    \label{tab:main}
    \resizebox{\textwidth}{!}{
    \begin{tabular}{lcc|cc|cc|cc|cc}
        \toprule
        \multirow{2}{*}{Model} & \multicolumn{2}{c}{FRWKV (Ours)} & \multicolumn{2}{c}{Leddam (2024)} & \multicolumn{2}{c}{Fredformer (2024)} & \multicolumn{2}{c}{iTrans. (2024)} & \multicolumn{2}{c}{TimeMixer (2024)} \\
        & MSE (Rank) & MAE (Rank) & MSE (Rank) & MAE (Rank) & MSE (Rank) & MAE (Rank) & MSE (Rank) & MAE (Rank) & MSE (Rank) & MAE (Rank) \\
        \midrule
        ETTm1 & 0.380 (1) & 0.385 (1) & 0.386 (4) & 0.397 (3) & 0.384 (3) & 0.395 (2) & 0.407 (8) & 0.410 (7) & 0.381 (2) & 0.395 (2) \\
        ETTm2 & 0.272 (1) & 0.316 (1) & 0.281 (4) & 0.325 (4) & 0.279 (3) & 0.324 (3) & 0.288 (6) & 0.332 (6) & 0.275 (2) & 0.323 (2) \\
        ETTh1 & 0.433 (2) & 0.430 (3) & 0.431 (1) & 0.429 (2) & 0.435 (3) & 0.426 (1) & 0.454 (5) & 0.447 (5) & 0.447 (4) & 0.440 (4) \\
        ETTh2 & 0.368 (3) & 0.391 (1) & 0.373 (4) & 0.399 (4) & 0.365 (2) & 0.393 (2) & 0.383 (5) & 0.407 (6) & 0.364 (1) & 0.395 (3) \\
        ECL   & 0.171 (2) & 0.257 (1) & 0.169 (1) & 0.263 (2) & 0.176 (3) & 0.269 (3) & 0.178 (4) & 0.270 (4) & 0.182 (5) & 0.272 (5) \\
        Exchange & 0.349 (2) & 0.396 (2) & 0.354 (3) & 0.402 (3) & 0.333 (1) & 0.391 (1) & 0.360 (5) & 0.403 (4) & 0.387 (6) & 0.416 (7) \\
        Weather & 0.243 (3) & 0.263 (1) & 0.242 (2) & 0.272 (3) & 0.246 (4) & 0.272 (3) & 0.258 (6) & 0.279 (4) & 0.240 (1) & 0.271 (2) \\
        Solar & 0.220 (2) & 0.220 (1) & 0.230 (4) & 0.264 (4) & 0.226 (3) & 0.262 (3) & 0.233 (5) & 0.262 (3) & 0.216 (1) & 0.280 (5) \\
        Avg. Rank & 2.00 (1) & 1.38 (1) & 2.88 (4) & 3.13 (3) & 2.75 (2) & 2.63 (2) & 5.50 (5) & 5.00 (5) & 2.75 (3) & 3.75 (4) \\
        \midrule
        \multirow{2}{*}{Model} & \multicolumn{2}{c}{FreTS (2024)} & \multicolumn{2}{c}{Crossfm. (2023)} & \multicolumn{2}{c}{TimesNet (2023)} & \multicolumn{2}{c}{PatchTST (2023)} & \multicolumn{2}{c}{DLinear (2023)} \\
        & MSE (Rank) & MAE (Rank) & MSE (Rank) & MAE (Rank) & MSE (Rank) & MAE (Rank) & MSE (Rank) & MAE (Rank) & MSE (Rank) & MAE (Rank) \\
        \midrule
        ETTm1 & 0.407 (8) & 0.415 (8) & 0.513 (9) & 0.496 (9) & 0.400 (6) & 0.406 (5) & 0.387 (5) & 0.400 (4) & 0.403 (7) & 0.407 (6) \\
        ETTm2 & 0.335 (8) & 0.379 (8) & 0.757 (10) & 0.610 (10) & 0.291 (7) & 0.333 (7) & 0.281 (5) & 0.326 (5) & 0.350 (9) & 0.401 (9) \\
        ETTh1 & 0.488 (9) & 0.474 (9) & 0.529 (10) & 0.522 (10) & 0.458 (7) & 0.450 (6) & 0.469 (8) & 0.454 (8) & 0.456 (6) & 0.452 (7) \\
        ETTh2 & 0.550 (8) & 0.515 (8) & 0.942 (10) & 0.684 (10) & 0.414 (7) & 0.427 (7) & 0.384 (6) & 0.405 (5) & 0.559 (9) & 0.515 (9) \\
        ECL & 0.202 (7) & 0.290 (6) & 0.244 (10) & 0.334 (10) & 0.192 (6) & 0.295 (7) & 0.208 (8) & 0.295 (8) & 0.212 (9) & 0.300 (9) \\
        Exchange & 0.416 (7) & 0.439 (8) & 0.940 (8) & 0.707 (10) & 0.416 (7) & 0.443 (9) & 0.367 (6) & 0.404 (5) & 0.354 (4) & 0.414 (6) \\
        Weather & 0.255 (5) & 0.298 (7) & 0.259 (7) & 0.315 (8) & 0.259 (7) & 0.287 (6) & 0.259 (7) & 0.281 (5) & 0.265 (8) & 0.317 (9) \\
        Solar & 0.226 (3) & 0.254 (2) & 0.641 (9) & 0.639 (9) & 0.301 (7) & 0.319 (7) & 0.270 (6) & 0.307 (6) & 0.330 (8) & 0.401 (8) \\
        Avg. Rank & 6.88 (8) & 8.00 (9) & 9.13 (10) & 9.50 (10) & 6.75 (7) & 6.75 (7) & 6.38 (6) & 5.75 (6) & 7.50 (9) & 7.88 (8) \\
        \bottomrule
    \end{tabular}}
\end{table*}

\subsection{Ablation Study}
We ablate frequency-domain processing (FR) and linear attention (LA) on ETTh1/ETTm1/Weather (Table~\ref{tab:ablation}). The full model $(\mathrm{FR} + \mathrm{LA})$ is best; removing FR or disabling LA consistently degrades accuracy. FR captures global periodicity and LA enables efficient state recursion—their combination is consistently superior and robust. The removed module is replaced with a single fully connected layer (linear projection) that preserves the input-output dimensionality and uses the same parameters, ensuring that any performance change is attributable to the ablated component itself.

\begin{table}[t]
    \centering
    \caption{Ablation on frequency-domain processing (FR) and linear attention (LA).}
    \label{tab:ablation}
    \begin{tabular}{cc|cc|cc|cc}
        \toprule
        FR & LA & \multicolumn{2}{c|}{ETTh1} & \multicolumn{2}{c|}{ETTm1} & \multicolumn{2}{c}{Weather} \\
        & & MSE & MAE & MSE & MAE & MSE & MAE \\
        \midrule
        \checkmark & \checkmark & 0.435 & 0.428 & 0.380 & 0.383 & 0.242 & 0.263 \\
        $\times$ & \checkmark & 0.460 & 0.451 & 0.411 & 0.407 & 0.254 & 0.274 \\
        \checkmark & $\times$ & 0.470 & 0.455 & 0.391 & 0.389 & 0.250 & 0.268 \\
        \bottomrule
    \end{tabular}
\end{table}

\section{CONCLUSION}
We propose FRWKV, a frequency-domain linear attention framework for long-term forecasting. Across 8 datasets, FRWKV achieves competitive performance, with ablations confirming the complementary benefits of frequency-domain processing and linear attention. FRWKV establishes a new paradigm for scalable time series modeling that bridges computational efficiency and predictive accuracy.

\bibliographystyle{unsrt}

\end{document}